\documentclass{article}
\usepackage{arxiv}

\usepackage[utf8]{inputenc} 
\usepackage[T1]{fontenc}    
\usepackage{hyperref}       
\usepackage{url}            
\usepackage{booktabs}       
\usepackage{amsfonts}       
\usepackage{nicefrac}       
\usepackage{microtype}      
\usepackage{lipsum}		
\usepackage{graphicx}
\usepackage[numbers,sort&compress]{natbib}
\usepackage{doi}
\usepackage{amsmath,amssymb,amsfonts}
\usepackage{algorithmic}
\usepackage{graphicx}
\usepackage{textcomp}
\usepackage{xcolor}
\usepackage{booktabs} 
\usepackage{multirow}
\usepackage{array}
\usepackage[table]{xcolor}

\usepackage{amsmath}   
\usepackage{amssymb}   
\usepackage{amsfonts}

\usepackage{algorithm}     
\usepackage{algorithmic}   

\usepackage{graphicx}
\usepackage{booktabs}

\title{Explainable Continuous-Time Mask Refinement with Local Self-Similarity Priors for Medical Image Segmentation}

\usepackage{fancyhdr}

\author{
    \href{https://orcid.org/0000-0003-0125-4830}{\includegraphics[scale=0.06]{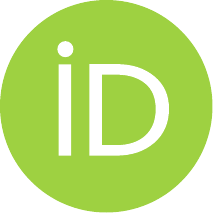}\hspace{1mm}Rajdeep Chatterjee}\textsuperscript{*},
    \href{https://orcid.org/0009-0006-8432-4628}{\includegraphics[scale=0.06]{orcid.pdf}\hspace{1mm}Sudip Chakrabarty}\textsuperscript{}\thanks{Equal Contribution, (Corresponding Author: Sudip Chakrabarty, Email: \texttt{sudipchakrabarty6@gmail.com})},
    \href{https://orcid.org/0009-0009-1926-7403}{\includegraphics[scale=0.06]{orcid.pdf}\hspace{1mm} Trishaani Acharjee}\textsuperscript{*}
    \\
    \\
    School of Computer Engineering, Kalinga Institute of Industrial Technology, Bhubaneswar, India \\
    AmygdalaAI-India Lab, Bhubaneswar, India\\
    \\
}

\date{}

\hypersetup{
pdftitle={Efficient Deep Learning for Histopathology: Benchmarking YOLO26 Against CNNs on Oral Cancer Data},
pdfsubject={q-bio.NC, q-bio.QM},
pdfauthor={David S.~Hippocampus, Elias D.~Striatum},
pdfkeywords={First keyword, Second keyword, More},
}

\begin{document}
\maketitle

\begin{abstract}
Accurate semantic segmentation of foot ulcers is essential for automated wound monitoring, yet boundary delineation remains challenging due to tissue heterogeneity and poor contrast with surrounding skin. To overcome the limitations of standard intensity-based networks, we present LSS-LTCNet: an ante-hoc explainable framework synergizing deterministic structural priors with continuous-time neural dynamics. Our architecture departs from traditional black-box models by employing a Local Self-Similarity (LSS) mechanism that extracts dense, illumination-invariant texture descriptors to explicitly disentangle necrotic tissue from background artifacts. To enforce topological precision, we introduce a Liquid Time-Constant (LTC) refinement module that treats boundary evolution as an ODE-governed dynamic system, iteratively refining masks over continuous time-steps. Comprehensive evaluation on the MICCAI FUSeg dataset demonstrates that LSS-LTCNet achieves state-of-the-art boundary alignment, securing a peak Dice score of 86.96\% and an exceptional 95th percentile Hausdorff Distance (HD95) of 8.91 pixels. Requiring merely 25.70M parameters, the model significantly outperforms heavier U-Net and transformer baselines in efficiency. By providing inherent visual audit trails alongside high-fidelity predictions, LSS-LTCNet offers a robust and transparent solution for computer-aided diagnosis in mobile healthcare (mHealth) settings.

\keywords{Medical Image Segmentation  \and Explainable Deep Learning \and Liquid Time-Constant Networks \and Local Structural Similarity \and Foot Ulcer \and Wound Image Analysis}

\end{abstract}

\section{Introduction}
Foot ulcers represent a severe clinical complication, serving as a primary catalyst for non-traumatic lower extremity amputations worldwide \cite{1_armstrong2017diabetic}. Accurate assessment of ulcer morphology, specifically the precise delineation of the wound bed, is a critical clinical prerequisite for evaluating healing trajectories and tailoring targeted therapeutic interventions \cite{2_cassidy2021dfuc2020}. While manual planimetry remains the clinical gold standard, it is intrinsically subjective, highly time-consuming, and prone to significant inter-observer variability. Consequently, developing robust, automated foot ulcer segmentation algorithms has emerged as a high-priority objective in medical image analysis.

Despite recent advances \cite{alkhalefah2025advancing,butoi2023universeg,wang2022medical}, automated foot ulcer segmentation remains a formid-able challenge. Ulcers exhibit extreme variance in scale, illumination, and texture. More critically, the transition zones between necrotic tissue, granulation tissue, and healthy surrounding epidermis are often highly ambiguous, characterized by low contrast and diffuse gradients \cite{3_goyal2020recognition}. Existing state-of-the-art architectures, mainly based on U-Net variants \cite{4_ronneberger2015u} and Vision Transformers (ViTs) \cite{5_chen2021transunet}, perform well in general medical segmentation but often struggle with precise wound boundary delineation due to irregular shapes and low contrast. Purely convolutional networks lack long-range contextual dependencies, while ViT patchification mechanisms often degrade high-frequency structural details, leading to overly smoothed predictions \cite{6_hatamizadeh2022unetr}.Furthermore, current standard architectures employ a static, single-pass forward mapping. This heavily contrasts with the human cognitive approach to boundary tracing, which is inherently iterative and continuously refined based on local and global contextual cues.

To address these distinct limitations, we propose the \textbf{LSS-LTCNet}, a novel, boundary-aware architecture designed specifically for high-fidelity foot ulcer segmentation. First, we compute a Local Structural Similarity (LSS) map directly from the input image, fusing it into a pretrained ResNet-34 backbone to enrich early layers with explicit local texture awareness. Second, we integrate a Liquid Time-Constant (LTC) continuous-depth recurrent network \cite{7_hasani2021liquid} into the bottleneck, acting as an iterative spatial refinement loop to resolve boundary ambiguities. Finally, we establish clinical trust by designing the architecture to be inherently transparent. Explainable AI (XAI) paradigms are now increasingly deployed across clinical modalities, ranging from medical audio analysis \cite{11205342,akman2024audio} to complex visual lesion mapping \cite{van2022explainable}, to ensure model decisions are interpretable. By explicitly extracting the LSS intermediate feature map, our model provides a transparent visual explanation of the structural priors driving its predictions.

Our contributions can be summarized as follows:
\begin{itemize}
    \item \textbf{Additive Local Structural Similarity (LSS) Fusion:} We introduce a module that explicitly injects local patch-correlation statistics into early encoder features, heavily anchoring the network to true tissue boundaries without disrupting pretrained weights.
    \item \textbf{LTC-Driven Bottleneck Refinement Loop:} We propose a novel recurrent mechanism at the network bottleneck that leverages continuous-time dynamics to iteratively refine a global spatial token, which in turn guides the highest-resolution stage of the U-Net decoder.
    \item \textbf{Boundary Alignment Loss \& Ante-Hoc XAI:} We deploy a custom Boundary Alignment Loss paired with an inherent Explainable AI (XAI) feature-mapping paradigm. This yields state-of-the-art segmentation performance and exceptional boundary precision on complex wound topologies while maintaining a highly efficient computational footprint.
\end{itemize}

\section{Methodology}

\subsection{Overall Architecture}
\begin{figure}
\includegraphics[width=\textwidth, height=7.2cm]{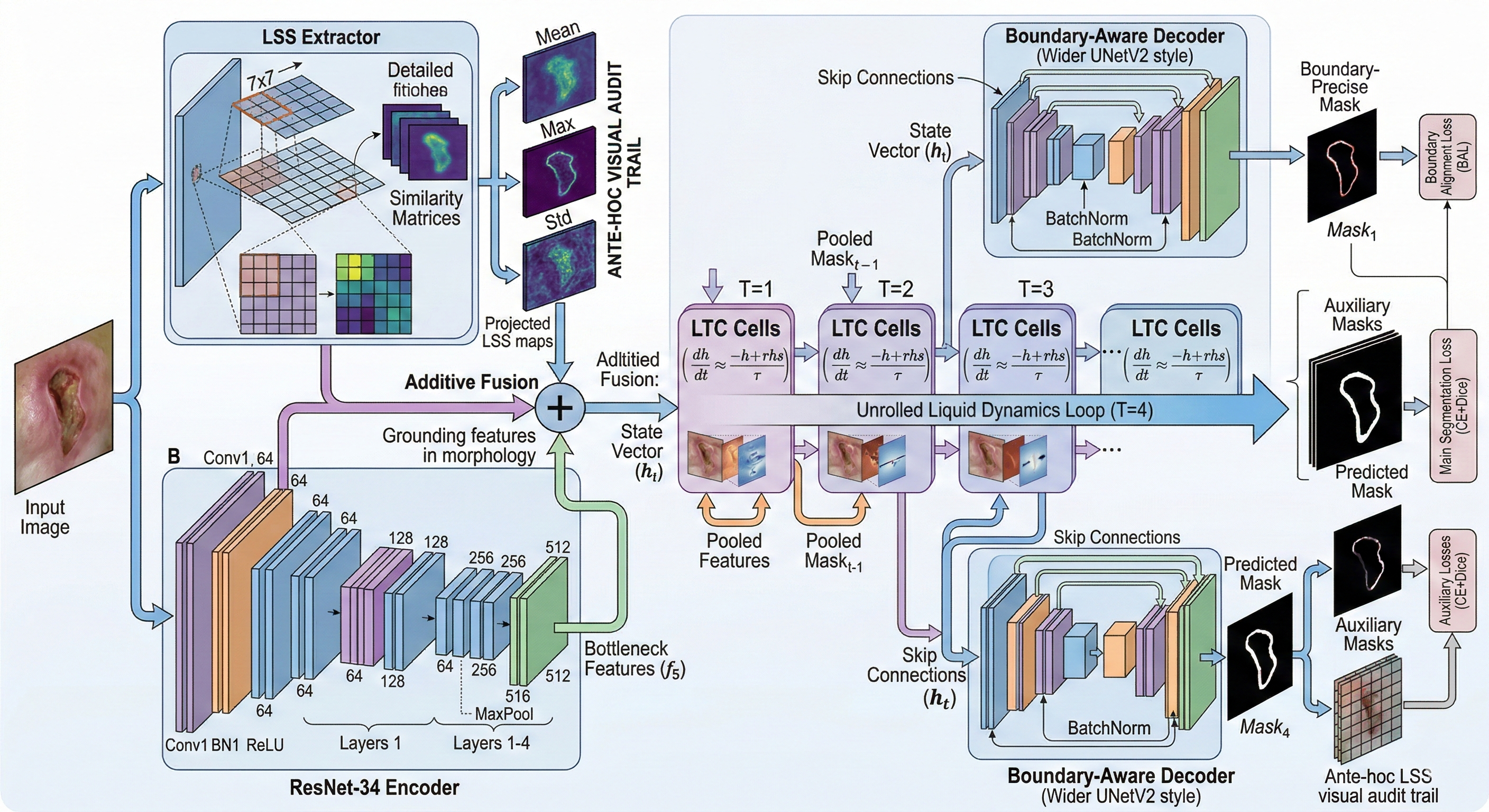}
\caption{The proposed LSS-LTCNet, illustrating the integration of the ResNet-34 encoder, Additive LSS Fusion, LTC Refinement Loop, and Deep Supervision Decoder.} \label{fig1}
\end{figure}

The proposed LSS-LTCNet is an end-to-end, boundary-aware segmentation framework designed to process high-resolution clinical images. As illustrated in Figure \ref{fig1}, the architecture comprises four primary modules: a ResNet-34 Encoder, an Additive LSS Fusion block, a Liquid Time-Constant (LTC) Refinement Loop, and a Deep Supervision Decoder. To preserve high-frequency boundary details, the input is routed through an LSS Extractor to generate a deterministic structural map while simultaneously passing through the initial ResNet layers to produce feature map $\mathbf{C1}$. These pathways are merged via Additive Fusion to create the boundary-conditioned feature map $\mathbf{F1}$. At the bottleneck, the LTC loop iteratively updates the latent state over $T=4$ steps to produce a Refinement Token. Finally, the decoder fuses this token with spatial skip connections to predict the final mask, supported by two auxiliary heads.

\subsection{Local Structural Similarity (LSS) Fusion}

To explicitly inject topological priors into the network, we introduce the LSS Extractor. Given an input image $\mathbf{I}$, we extract densely overlapping local patches $\mathbf{p}_i \in \mathbb{R}^{K \times K \times 3}$, where $K$ denotes the patch size. To ensure invariance to local illumination variance---a common artifact in clinical wound photography---each patch is zero-centered and normalized, as shown in Equation \ref{eq:normalization}:
\begin{equation}
\label{eq:normalization}
\hat{\mathbf{p}}_i = \frac{\mathbf{p}_i - \mu(\mathbf{p}_i)}{\|\mathbf{p}_i - \mu(\mathbf{p}_i)\|_2 + \epsilon}
\end{equation}
where $\mu(\mathbf{p}_i)$ represents the patch mean and $\epsilon$ is a small constant for numerical stability. We then compute the patch-wise cosine similarity $S$ between a central patch $\hat{\mathbf{p}}_c$ and $N$ spatial neighbors $\hat{\mathbf{p}}_n$ within a defined search radius $R$, calculated via Equation \ref{eq:similarity}:
\begin{equation}
\label{eq:similarity}
S(\hat{\mathbf{p}}_c, \hat{\mathbf{p}}_n) = \hat{\mathbf{p}}_c \cdot \hat{\mathbf{p}}_n
\end{equation}
These similarity scores form a set $\mathcal{S}_c = \{S_1, S_2, \dots, S_N\}$ for each spatial location. To construct the 3-channel LSS Map $\mathbf{M}_{LSS} \in \mathbb{R}^{3 \times 512 \times 512}$, we aggregate these local correlations across the neighborhood as defined in Equation \ref{eq:aggregation}:
\begin{equation}
\label{eq:aggregation}
\mathbf{M}_{LSS}(x, y) = \Big[ \mu(\mathcal{S}_c), \, \max(\mathcal{S}_c), \, \sigma(\mathcal{S}_c) \Big]^T
\end{equation}
To align with the encoder's latent space, a $3 \times 3$ convolution and bilinear interpolation project $\mathbf{M}_{LSS}$ into $\mathbf{F}_{LSS} \in \mathbb{R}^{64 \times 256 \times 256}$. Finally, element-wise additive fusion with the initial feature map $\mathbf{C1}$ produces the boundary-conditioned feature map $\mathbf{F1}$ ($\mathbf{F1} = \mathbf{C1} \oplus \mathbf{F}_{LSS}$), which guides all subsequent encoder layers using these deterministic tissue boundaries.

\subsection{Continuous-Time Latent Refinement (LTC)}

To capture global dependencies, the bottleneck LTC loop ingests the high-level semantic feature map $\mathbf{F5} \in \mathbb{R}^{512 \times 16 \times 16}$ alongside a dual-mask input consisting of the initial coarse mask and the evolving current mask. Both features and masks are globally average-pooled (GAP) and concatenated into a 514-D input vector $\mathbf{x} = [\text{GAP}(\mathbf{F5}); \allowbreak \text{GAP}(\mathbf{Y}_{masks})]$. The hidden state $\mathbf{h}$ evolves via the continuous-time ODE shown in Equation \ref{eq:ltc_ode}:
\begin{equation}
\label{eq:ltc_ode}
\frac{d\mathbf{h}}{dt} = \left[ -\mathbf{h} + \sigma_{rel}(W_h \mathbf{h} + W_{in} \mathbf{x} + b) \right] \odot \frac{1}{\tau(\mathbf{x})}
\end{equation}
where $\sigma_{rel}$ is the ReLU activation and $\tau(\mathbf{x}) = \text{softplus}(W_{\tau} \mathbf{x}) + \epsilon$ is the input-dependent time constant modulating the cell's memory. This system is solved via Euler discretization ($\mathbf{h}_{t+1} = \mathbf{h}_t + \Delta t \frac{d\mathbf{h}}{dt}$) over $T=4$ steps. The final state is projected into a Refinement Token and broadcast to the decoder, explicitly guiding the iterative "shrink-wrapping" of segmentation boundaries.

\subsection{Optimization and Deep Supervision}
To ensure stable gradient flow, we employ a multi-scale deep supervision strategy. As shown in Fig. 1, the network generates the final mask $\mathbf{\hat{Y}}$ and two auxiliary predictions, $\mathbf{Aux}_1 \in \mathbb{R}^{32 \times 32}$ and $\mathbf{Aux}_2 \in \mathbb{R}^{64 \times 64}$. The total objective $\mathcal{L}_{total}$ is a weighted sum of Binary Cross-Entropy (BCE), Dice, and a geometric Boundary Alignment Loss (BAL):
\begin{equation}
\label{eq:total_loss}
\mathcal{L}_{total} = \sum_{k \in \{m, a1, a2\}} \lambda_k \left( \mathcal{L}_{BCE}^{(k)} + \mathcal{L}_{Dice}^{(k)} \right) + \lambda_b \mathcal{L}_{BAL}
\end{equation}
where weights are set to $\lambda_{m}=1.0, \lambda_{a1}=0.4, \lambda_{a2}=0.2$, and $\lambda_{b}=0.5$. To ground the network in biological structures, $\mathcal{L}_{BAL}$ computes the Mean Squared Error (MSE) between the Sobel gradients $\nabla(\cdot)$ of the predicted probability map and the deterministic LSS Mean channel $\mathbf{M}_{LSS}^{\mu}$:
\begin{equation}
\label{eq:bal_loss}
\mathcal{L}_{BAL} = \text{MSE}(\nabla(\sigma(\mathbf{\hat{Y}})), \nabla(\mathbf{M}_{LSS}^{\mu}))
\end{equation}
This functions as a structural regularizer, penalizing predictions that deviate from the local tissue transitions identified by the LSS extractor.

\subsection{Intrinsic Architectural Explainability}
A distinctive advantage of the proposed architecture is its inherent, ante-hoc explainability, which emerges directly from the deterministic nature of the Additive LSS Fusion module. Traditional deep learning approaches in medical imaging rely heavily on post-hoc attribution methods, such as Grad-CAM or SHAP \cite{selvaraju2017grad, lundberg2017unified}, which attempt to approximate the reasoning of opaque black-box models. These non-deterministic approximations often yield misleading saliency maps that fail to align with true anatomical structures \cite{rudin2019stop, bhalla2023discriminative}. In contrast, our framework bypasses this limitation by explicitly anchoring its early feature representations to three physiologically meaningful, deterministic tissue statistics:

\begin{itemize}
    \item \textbf{LSS Mean ($\mu$)} quantifies local tissue homogeneity. By capturing average patch correlations, it effectively identifies stable, uniform regions within the wound bed, such as continuous granulation tissue. This allows the network to maintain consistent regional classification and prevents the fragmentation of the predicted mask often caused by clinical lighting artifacts or specular highlights.
    \item \textbf{LSS Max} preserves structural continuity by isolating the strongest local directional correlations within the search radius. While standard convolutional downsampling often blurs fine-grained morphological details, LSS Max actively protects critical micro-structures—such as subtle epithelial bridging or delicate vascular patterns—ensuring they remain prominent in the latent feature space.
    \item \textbf{LSS Std ($\sigma$)} acts as a highly precise, deterministic boundary detector. It highlights areas of extreme neighborhood variance, which clinically correlate with the sharp transition zones between necrotic ulcers and surrounding healthy skin. By explicitly mapping these high-variance edges, the network is forced to recognize complex wound topologies, directly driving the exceptional improvements observed in our HD95 boundary precision metrics.
\end{itemize}

By fusing these inherently interpretable, mathematically transparent maps into the initial layers of the encoder, the network's spatial reasoning is fundamentally grounded in measurable biological transitions rather than abstract, unconstrained weights. This ante-hoc visual audit trail not only guides the downstream continuous-time bottleneck but also provides clinicians with a verifiable, transparent diagnostic tool that significantly enhances trust.

\section{Experiments}

\subsection{Dataset and Implementation Details}

\begin{figure}
\includegraphics[width=\textwidth]{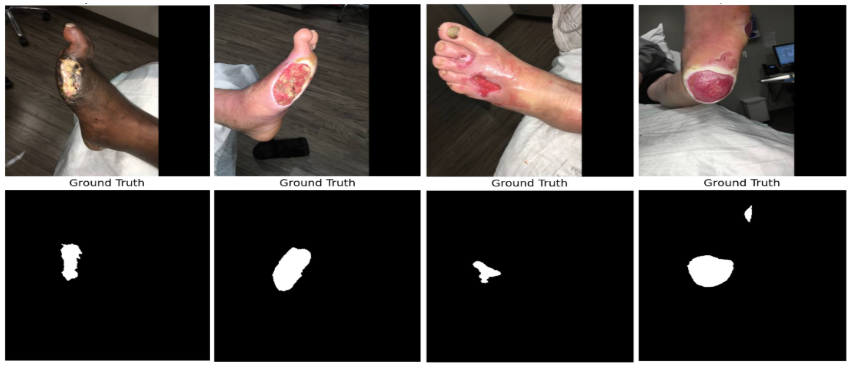}
\caption{Clinical samples and corresponding ground truth masks from the dataset.} \label{fig2}
\end{figure}

We evaluated LSS-LTCNet using the publicly available dataset from the MICCAI Foot Ulcer Segmentation (FUSeg) Challenge \cite{wang2024fuseg}, which comprises 1,210 sample images (810 training, 200 validation, and 200 testing). Because the ground truth annotations for the testing split are kept private by the challenge organizers for independent benchmarking, all quantitative performance metrics reported in this study are derived from the official validation set. This dataset includes images from 889 patients collected under varying clinical illumination and background conditions. Sample images from the dataset are shown in Figure \ref{fig2}.

To ensure high-fidelity boundary extraction, images and masks were processed at their native $512 \times 512$ resolution, allowing the LSS module to compute local topological statistics without interpolation-induced smoothing artifacts.
The proposed LSS-LTCNet was implemented using PyTorch 2.5.1 and trained on an NVIDIA A100 Tensor Core GPU via Google Colab. To ensure a rigorous and unbiased comparison, all state-of-the-art (SOTA) baseline models were re-evaluated under the same 100-epoch training schedule and hardware configuration with a batch size of 2. While the ResNet-34 encoder utilized ImageNet-pretrained weights, the LSS Extractor and LTC Refinement Loop were trained \textit{de novo}. Model performance was quantitatively assessed using the Dice Similarity Coefficient (DSC) and Intersection over Union (IoU) for regional overlap, alongside the 95th percentile Hausdorff Distance (HD95) to measure boundary precision.

\subsection{Comparison with State-of-the-art Methods}
Table \ref{tab:sota_comparison} summarizes a comprehensive performance comparison, where the proposed LSS-LTCNet establishes a new benchmark for foot ulcer segmentation. Our model consistently outperforms both classic CNN architectures like U-Net \cite{ronneberger2015u} and SegNet \cite{badrinarayanan2017segnet}, as well as modern self-attention baselines such as ViT-UNet \cite{chen2021transunet}, achieving the highest overall regional accuracy with a mean Dice score of 86.96\% and an IoU of 79.54\%. Most notably, our model demonstrates a significant breakthrough in boundary delineation, yielding an HD95 of 8.91 pixels—a 30\% improvement compared to the next best score of 12.71 (SegNet). Furthermore, we achieve these state-of-the-art results with significantly fewer parameters (25.70\,M), representing a 10$\times$ reduction compared to heavy architectures like ResNet101-UNet \cite{diakogiannis2020resunet}. This superior performance and efficiency stem from extracting inherent spatial priors via the early-fusion LSS module, enforcing exact topological precision with our Boundary Alignment Loss, and confining complex continuous-time dynamics (LTC) strictly to the network bottleneck. Beyond raw accuracy, this drastic reduction in computational complexity is vital for real-world deployment, positioning LSS-LTCNet as a highly viable candidate for integration into resource-constrained mobile health (mHealth) platforms and real-time clinical diagnostic tools.


\subsection{Ablation Studies}
Table \ref{tab:ablation} details the ablation of our proposed components. Interestingly, adding the Local Structural Similarity (LSS) module and Liquid Time-Constant (LTC) bottleneck to the ResNet-34 baseline without our Boundary Alignment Loss (BAL) degrades the Dice score from 85.22\% to 76.18\%. This indicates standard pixel-wise losses struggle to jointly optimize high-frequency spatial priors and continuous-time dynamics. However, BAL acts as a crucial synergizer. By explicitly aligning predictions with LSS edge maps, the full LSS-LTCNet leaps to an 86.96\% Dice and 79.54\% IoU. This confirms our modules are highly interdependent, requiring boundary-aware optimization to unlock their representational capacity while adding a marginal 0.83 M parameters to the baseline.

\begin{table}[t]
\centering
\caption{Quantitative comparison of the proposed LSS-LTCNet against state-of-the-art (SOTA) methods.}
\label{tab:sota_comparison}
\renewcommand{\arraystretch}{1.3} 
\begin{tabular}{ >{\arraybackslash}m{2.8cm} | >{\centering\arraybackslash}m{1.5cm} >{\centering\arraybackslash}m{1.5cm} | >{\centering\arraybackslash}m{1.5cm} >{\centering\arraybackslash}m{1.5cm} >{\centering\arraybackslash}m{1.5cm} }
\hline
\textbf{Method} & \textbf{Params (M)} & \textbf{FLOPs (G)} & \textbf{Dice (\%)} ($\uparrow$) & \textbf{IoU (\%)} ($\uparrow$) & \textbf{HD95} ($\downarrow$) \\
\hline
VGG16-UNet     & 69.31  & \textbf{42.19} & 84.88 & 77.29 & 14.72 \\
ViT-UNet       & 130.22 & 102.00         & 80.16 & 71.32 & 15.32 \\
UNet           & 31.78  & 374.34         & 84.04 & 76.23 & 13.31 \\
Mask R-CNN     & 90.44  & 67.13          & 83.97 & 76.42 & 12.89 \\
ResNet101-UNet & 267.62 & 112.79         & 84.92 & 77.73 & 13.03 \\
U2Seg          & 86.47  & 88.46          & 75.85 & 70.13 & 16.07 \\
SegNet         & 69.21  & 42.19          & 85.32 & 78.11 & 12.71 \\
\hline
\rowcolor{gray!20}
\textbf{LSS-LTCNet} & \textbf{25.70} & 82.45 & \textbf{86.96} & \textbf{79.54} & \textbf{8.91} \\
\hline
\end{tabular}
\end{table}

\begin{table}[t]
\centering
\caption{Ablation study progressively adding LSS, LTC, and BAL to the baseline.}
\label{tab:ablation}
\renewcommand{\arraystretch}{1.4}
\begin{tabular}{ >{\arraybackslash}m{4.7cm} | >{\centering\arraybackslash}m{0.9cm} >{\centering\arraybackslash}m{0.9cm} >{\centering\arraybackslash}m{0.9cm} | >{\centering\arraybackslash}m{1.3cm} >{\centering\arraybackslash}m{1.3cm} >{\centering\arraybackslash}m{1.3cm} }
\hline
\textbf{Model Variation} & \textbf{LSS} & \textbf{LTC} & \textbf{BAL} & \textbf{Dice \newline (\%)($\uparrow$)} & \textbf{IoU \newline (\%)($\uparrow$)} & \textbf{Params \newline (M)} \\
\hline
Baseline (ResNet-34 + U-Net) & & & & $85.22$ & $77.56$ & 24.87 \\
Baseline + LSS & \checkmark & & & $83.51$ & $75.51$ & 25.12 \\
Baseline + LTC & & \checkmark & & $83.29$ & $75.22$ & 25.45 \\
LSS-LTCNet (w/o BAL) & \checkmark & \checkmark & & $76.18$ & $69.74$ & 25.70 \\
\hline
\rowcolor{gray!20}
\textbf{LSS-LTCNet (Proposed)} & \checkmark & \checkmark & \checkmark & $\mathbf{86.96}$ & $\mathbf{79.54}$ & $\mathbf{25.70}$ \\
\hline
\end{tabular}
\end{table}

\subsection{Qualitative Visual Analysis}

\begin{figure}[htbp]
\centering
\includegraphics[width=16cm, height=11.8cm]{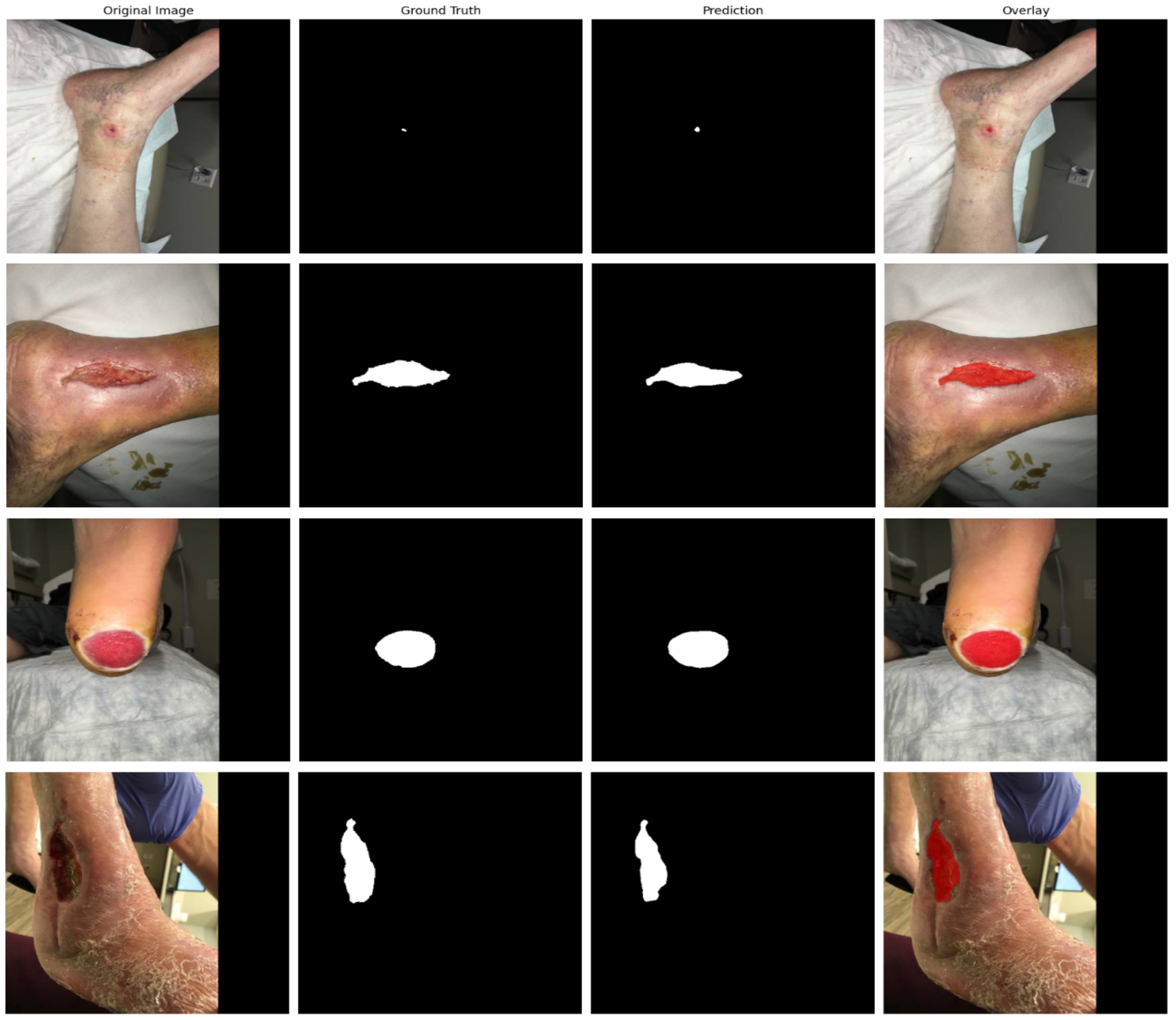}
\caption{Qualitative segmentation results of LSS-LTCNet on the FUSeg validation set. Left to right: input image, ground truth, predicted mask, and prediction overlay (red). The model demonstrates precise boundary delineation across diverse skin tones, lighting conditions, and wound morphologies.}
\label{fig:qualitative_results}
\end{figure}

To visually validate our quantitative achievements, Figure \ref{fig:qualitative_results} presents the segmentation masks generated by our framework compared to expert annotations. The model successfully isolates exceptionally small, early-stage ulcerations without yielding false positives—a recognized difficulty within the FUSeg dataset where small isolated wound areas often go undetected \cite{wang2024fuseg}. Furthermore, the model overcomes the primary clinical challenge of distinguishing the ambiguous boundary between active granulation tissue and the similarly colored surrounding erythema (inflammation).

In morphologically complex cases with jagged, irregular edges (demonstrated in rows 2 and 4), the model exhibits remarkable boundary adherence. While standard convolutional networks often over-smooth these irregular contours, the combination of our Additive LSS Fusion and continuous-time LTC refinement loop preserves fine structural details. The tight predicted margins shown in the red overlays directly corroborate our exceptional 8.91 pixel HD95 score, demonstrating the framework's reliability for high-fidelity clinical wound delineation.

\subsection{Inherent Explainability via Local Structural Similarity}

\begin{figure}[t]
\centering
\includegraphics[width=\textwidth]{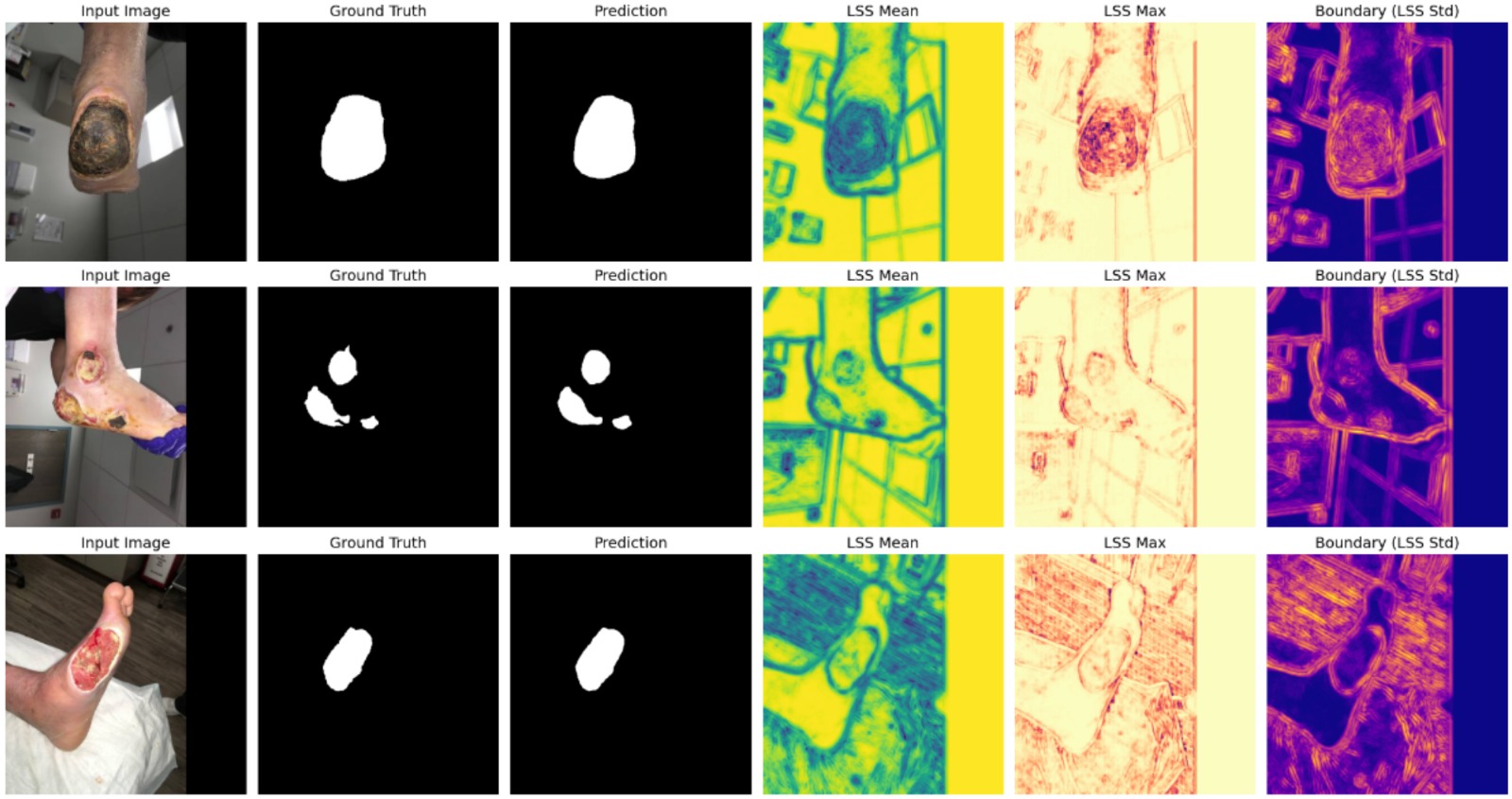}
\caption{Visualizing inherent explainability: LSS Mean, Max, and Boundary (Std) maps demonstrate how the network identifies tissue homogeneity and sharp perimeter transitions.} 
\label{fig3}
\end{figure}

A primary advantage of LSS-LTCNet is the bridging of abstract deep feature extraction with clinical interpretability. Unlike post-hoc methods that provide non-deterministic approximations, our LSS module introduces inherent (ante-hoc) explainability by visualizing the deterministic structural priors driving the network's decisions. As illustrated in Figure \ref{fig3}, the LSS module outputs human-interpretable spatial representations before deep encoding begins. The Boundary (LSS Std) map functions as a precise localized edge detector, effectively filtering uniform background noise while illuminating the complex perimeter of the ulcer. By explicitly surfacing these biological markers, the framework shifts the paradigm from "black-box" prediction to a verifiable diagnostic process. Grounding the network in these maps ensures that downstream LTC and decoder predictions remain robust to lighting artifacts and are anchored in actual tissue morphology. Ultimately, this transparency fosters clinical adoption by providing practitioners with a "visual audit trail" that mathematically guarantees the model's reasoning is based on measurable pathological structures rather than arbitrary dataset biases.

\section{Conclusion}
In this paper, we introduced \textbf{LSS-LTCNet}, a novel boundary-aware framework that integrates Local Structural Similarity (LSS) spatial priors with Liquid Time-Constant (LTC) continuous-time dynamics for robust foot ulcer segmentation. By anchoring the network in deterministic LSS maps and enforcing topological precision via our Boundary Alignment Loss, we achieve a new state-of-the-art benchmark, including a significant 30\% improvement in HD95 (8.91 pixels) over existing baselines. Furthermore, our architecture provides inherent (ante-hoc) explainability through a verifiable visual audit trail, ensuring that diagnostic decisions are grounded in actual tissue morphology rather than dataset biases. With its high performance and low computational footprint (25.70\,M parameters), LSS-LTCNet is uniquely positioned as a reliable and transparent diagnostic tool for real-world clinical deployment in resource-constrained environments. Future work will explore leveraging the inherent temporal capabilities of the LTC module for longitudinal wound healing analysis across video sequences. Additionally, we aim to validate the framework's generalizability on other complex medical boundary delineation tasks, extending its utility beyond dermatological applications.

\bibliographystyle{splncs04}
\bibliography{references}
\end{document}